\PassOptionsToPackage{table}{xcolor}
\documentclass[sigconf,screen,nonacm]{acmart}

\setcopyright{none}
\usepackage{amsmath}
\usepackage{array}
\usepackage{multirow}
\usepackage{pifont}
\usepackage{placeins}
\usepackage{enumitem}
\usepackage{hyperref}

\definecolor{lightblue}{rgb}{0.9, 0.95, 1.0}
\definecolor{lightpurple}{rgb}{0.83, 1, 0.73}
\definecolor{mygreen}{RGB}{0,153,0}
\definecolor{myred}{RGB}{204,0,0}
\definecolor{rowblue}{RGB}{235,245,250}

\newcommand{\cmark}{\raisebox{-0.15ex}{\scalebox{1.15}{\textcolor{mygreen}{\ding{51}}}}}
\newcommand{\xmark}{\raisebox{-0.15ex}{\scalebox{1.15}{\textcolor{myred}{\ding{55}}}}}

\title{MedXplore: Towards Reliable and Unbiased Generalized Category Discovery in Medical Imaging}

\author{Jianwei He}
\authornote{Jianwei He and Kailin Lyu contributed equally to this work.}
\affiliation{%
  \institution{Institute of Automation, Chinese Academy of Sciences}
  \city{Beijing}
  \country{China}
}

\author{Kailin Lyu}
\authornotemark[1]
\affiliation{%
  \institution{Institute of Automation, Chinese Academy of Sciences}
  \city{Beijing}
  \country{China}
}

\author{Junhao Dong}
\affiliation{%
  \institution{Nanyang Technological University}
  \city{Singapore}
  \country{Singapore}
}

\author{Long Xiao}
\affiliation{%
  \institution{Institute of Automation, Chinese Academy of Sciences}
  \city{Beijing}
  \country{China}
}

\author{Wenjie Hou}
\affiliation{%
  \institution{Institute of Automation, Chinese Academy of Sciences}
  \city{Beijing}
  \country{China}
}

\author{Jingze Lu}
\affiliation{%
  \institution{Institute of Automation, Chinese Academy of Sciences}
  \city{Beijing}
  \country{China}
}

\author{Di Wu}
\affiliation{%
  \institution{Institute of Automation, Chinese Academy of Sciences}
  \city{Beijing}
  \country{China}
}

\author{Lin Shu}
\affiliation{%
  \institution{Institute of Automation, Chinese Academy of Sciences}
  \city{Beijing}
  \country{China}
}

\author{Jie Hao}
\authornote{Corresponding author.}
\affiliation{%
  \institution{Institute of Automation, Chinese Academy of Sciences}
  \city{Beijing}
  \country{China}
}
\email{jie.hao@ia.ac.cn}

\renewcommand{\shortauthors}{He et al.}



\begin{document}

\begin{abstract}
Deep learning has shown strong potential in medical image analysis, but most existing methods rely on large-scale annotations and a closed-world assumption that rarely holds in clinical practice. Although Generalized Category Discovery (GCD) has advanced rapidly on natural images, it remains underexplored in medical imaging. To address this issue, we propose MedXplore, a unified framework for reliable and unbiased medical GCD, optimizing from both perceptual and decision levels. Specifically, at the perceptual level, taking a frequency domain perspective, Frequency-SNR Adaptive Attention and Consistency (FAAC) performs learnable full-spectrum filtering and global-local energy contrast activation to not only highlight local abnormal signals relative to the global context, but also provide reliable semantic anchors for patch consistency learning. At the decision level, Adaptive Cosine-Angular Margin (ACAM) adjusts angular margins using semantic difficulty and feature confidence to balance intra-class compactness and inter-class separability. Together, the two modules improve lesion-sensitive representation learning and mitigate old-class bias. Experiments on multiple benchmarks show an average \textbf{8.5\%} gain in \textit{All} accuracy over the strongest competing methods. On Kvasir, MedXplore reduces false-old errors from 14.50\% to 0.80\%, demonstrating strong robustness under severe old-new ambiguity.

\end{abstract}

\maketitle

\section{Introduction}
\label{sec:intro}

Deep learning-based recognition models have advanced rapidly~\cite{he2016deep, li2022challenging, vaswani2017attention, dosovitskiy2020image}, but most still rely on large annotated datasets and a closed-world assumption, namely that unlabeled samples belong to the same categories as labeled training data. This assumption breaks down in real-world medical settings, where exhaustive disease coverage is infeasible and new or rare conditions continue to emerge~\cite{vaze2022opensetrecognitiongoodclosedset, geng2020recent, han2021autonovel, lee2022deep, roberts2021common}. Developing models that can recognize known diseases while reliably discovering unknown ones is therefore a central challenge.

\begin{figure*}[t]
    \centering
    \includegraphics[width=0.9\textwidth]{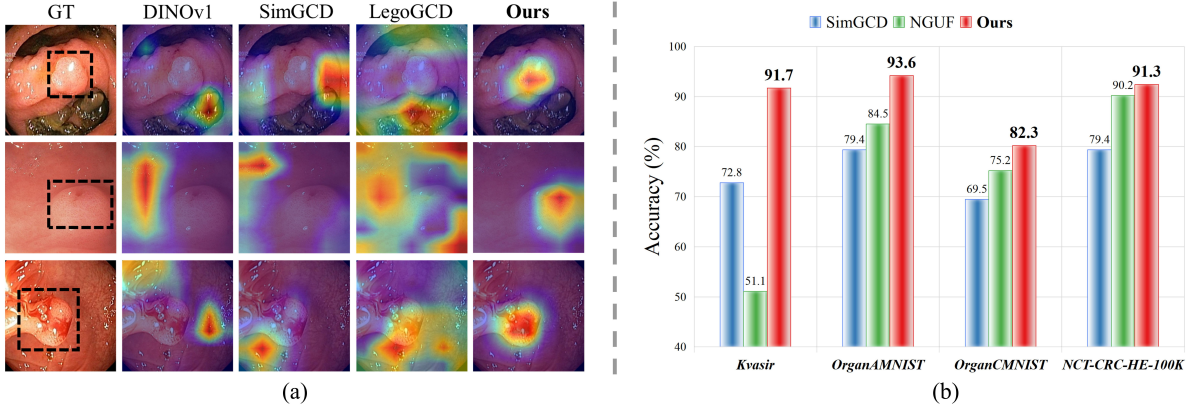}
    \caption{\textbf{Qualitative and quantitative comparison of MedXplore (Ours) against state-of-the-art (SOTA) methods.}
    \textbf{(a)} Attention visualizations on samples from the Kvasir-seg~\cite{jha2020kvasir} dataset. Compared to DINOv1~\cite{caron2021emerging}, SimGCD~\cite{wen2023parametric}, and LegoGCD~\cite{cao2024solving}, which exhibit attention drift or diffuse localization, our method accurately focuses on discriminative lesion regions, showing high alignment with the ground truth.
    \textbf{(b)} Comparison of `All' classes accuracy on multiple benchmark datasets.}
    \Description{A two-part figure comparing MedXplore with prior methods. Part (a) shows several medical image examples with attention maps from DINOv1, SimGCD, LegoGCD, MedXplore, and ground truth masks, where MedXplore aligns more closely with lesion regions. Part (b) shows a quantitative comparison of all-class accuracy across multiple benchmark datasets, with MedXplore outperforming competing methods.}
    \label{Absfig}
\end{figure*}

Although Generalized Category Discovery (GCD) has made substantial progress on natural-image benchmarks~\cite{vaze2022generalized, choi2024contrastive, wen2023parametric, cao2024solving}, it remains underexplored in medical imaging. Unlike natural images, medical images often contain subtle lesion cues and weaker semantic separation, causing existing GCD methods to suffer from attention drift and poor clustering when transferred to this domain (Fig.~\ref{Absfig}(a)). This raises a key question: \textit{\textbf{How can we achieve reliable identification and unbiased discovery of unknown disease categories in medical imaging?}}

The gap mainly stems from two properties of medical images. \textbf{\textit{(i)}} Medical data show strong domain specificity and cross-device variability, including different modalities, scanning parameters, and tissue contrasts~\cite{midya2018influence, lambin2012radiomics, james2014medical, song2023radiomics}. Backbones pretrained on natural images~\cite{caron2021emerging, wen2023parametric} often fail to capture such modality-specific cues and structural priors, leading to weak lesion representations. Related multimodal studies likewise highlight the need for robust representation learning under heterogeneous sensory inputs and open-set uncertainty~\cite{lyu2026touchformer,xiao2026tacexpert}. \textbf{\textit{(ii)}} Medical images also exhibit high heterogeneity and imbalanced semantic granularity: many disease subtypes differ only in subtle local patterns~\cite{wei2021fine, yu2021benchmark}. Limited labeled data, due to privacy constraints and annotation cost~\cite{guan2024federated,shurrab2022self}, further amplifies old-class bias~\cite{vaze2022generalized, wen2023parametric}, yielding blurred decision boundaries, cluster drift, and unstable representations. This effect is especially severe on endoscopic data such as Kvasir~\cite{pogorelov2017kvasir}, where old and novel classes may share nearly identical global appearance while differing only in subtle local mucosal patterns (see Fig.~\ref{fig:kvasir_bias}). Since such lesion evidence often appears as subtle local deviations, exploiting frequency-domain cues emerges as a natural direction to capture these anomalies.

To address these challenges, we propose \textbf{MedXplore}, a medical GCD framework that jointly models perceptual and discriminative cues. At the perceptual level, Frequency-SNR Adaptive Attention and Consistency (FAAC) performs learnable full-spectrum filtering followed by global-local energy contrast activation, preserving both low-frequency structural cues and high-frequency boundary evidence while producing semantic anchors for Top-K patch selection and patch consistency. At the decision level, Adaptive Cosine-Angular Margin (ACAM) introduces adaptive angular and cosine margins to sharpen class boundaries according to sample difficulty and confidence. By providing reliable, less biased semantic anchors and high-quality similarity/confidence signals, FAAC enables ACAM to calibrate decision boundaries more effectively. Together, this perception-discrimination coupling encourages lesion-sensitive representations and more reliable discovery of unknown disease categories.

Our main contributions can be summarized as follows:
\begin{enumerate}[leftmargin=*, itemsep=1pt, topsep=1pt, parsep=0pt, partopsep=0pt]
    \item We propose \textbf{MedXplore}, a framework for unbiased and reliable generalized category discovery in medical imaging.
    \item We introduce \textbf{FAAC} and \textbf{ACAM} to jointly model perceptual and discriminative cues. FAAC performs dynamic frequency purification and statistics-based semantic anchor extraction, while ACAM adaptively balances intra-class compactness and inter-class separability.
    \item Extensive experiments demonstrate the effectiveness of MedXplore, achieving state-of-the-art performance across multiple medical datasets, including a remarkable \textbf{16.5\%} improvement on Kvasir over the strongest competitor. We further provide an explicit bias analysis showing that MedXplore sharply reduces false-old predictions under severe old--new ambiguity.
\end{enumerate}

\section{Related Work}
\label{sec:Related Work}

\subsection{Novel Class Discovery}
Novel Class Discovery (NCD) seeks to identify new classes within unlabeled data by leveraging knowledge from labeled classes~\cite{han2019learning, hsu2017learning}. Early approaches framed this problem as deep transfer clustering, where transferable representations from labeled classes guide the organization of unlabeled samples. AutoNovel, a pioneering effort in this field, uses a three-stage pipeline: pretraining, fine-tuning on the labeled set, and knowledge transfer based on ranking statistics to aggregate novel classes without semantic priors~\cite{han2021autonovel}. Subsequent research shifted towards unified objectives and robust pseudo-label construction. UNO introduced a unified training objective, combined with a prediction-swapping mechanism for pseudo-label assignment~\cite{fini2021unified}. OpenMix generates virtual samples by applying MixUp between labeled and unlabeled instances, improving robustness to noisy labels~\cite{zhong2021openmix}. NCL utilizes local neighborhood relations to aggregate pseudo-positive pairs, improving intra-class consistency and inter-class separability without relying on external semantics~\cite{zhong2021neighborhood}. Traditional NCD assumes a complete separation between labeled and unlabeled class spaces, which limits its applicability in complex real-world scenarios. Thus, we focus on GCD to explore mechanisms for more reliable and robust class recognition and discovery under open-world conditions.

\subsection{Generalized Category Discovery}
The core objective of generalized category discovery (GCD) is to model and distinguish both base and novel classes within unlabeled data, given a limited set of labeled base classes, thus simulating open-world recognition. Early methods often relied on self-supervised pretrained features, such as DINO-based representations, combined with contrastive learning and semi-supervised k-means clustering. However, these approaches were computationally intensive and inefficient~\cite{vaze2022generalized}. Later, parameterized frameworks replaced clustering with a classification head and integrated pseudo-label generation and entropy regularization, leading to significant improvements in efficiency and robustness, gradually becoming the dominant paradigm in GCD~\cite{wen2023parametric}. Building on this foundation, subsequent research advanced feature extraction and pseudo-labeling, achieving continuous progress on natural image benchmarks. For example, LegoGCD addressed catastrophic forgetting in SimGCD by introducing regularization for potential known class samples~\cite{cao2024solving}, SPTNet enhanced discriminative capability through spatial prompt tuning~\cite{wang2024sptnet}, and ClearGCD explicitly mitigated shortcut learning to improve robust category discovery~\cite{lyu2026cleargcd}. Recent work has explored GCD applications in medical imaging, but these methods primarily focus on prototype-level performance, often neglecting lesion morphology, tissue structure, and pathological heterogeneity, which limits further gains~\cite{feng2025neighbor, das2025medgcd}. Motivated by the structural and semantic properties of medical images, this paper proposes a GCD framework that improves reliable recognition and unbiased discovery of disease categories.

\section{Preliminaries}
\label{Preliminaries}

\subsection{Problem Formulation}
\label{Problem Formulation}

We consider data arising from an open-world setting, consisting of both labeled and unlabeled samples. Let the entire dataset be denoted as $D = D_l \cup D_u$. The labeled subset is defined as $D_l = \{(x_i^l, y_i^l)\}_{i=1}^{n_l} \subset \mathcal{X} \times Y_l$, where $Y_l$ represents the set of known categories. The unlabeled subset is given by $D_u = \{x_j^u\}_{j=1}^{n_u} \subset \mathcal{X}$, with its latent label space $Y_u$ covering all categories such that $Y_l \subseteq Y_u$. We partition $Y_u$ into two disjoint subsets: old classes and novel classes, i.e., $Y_u = C_{\text{old}} \cup C_{\text{new}}, \; C_{\text{old}} = Y_l, \; C_{\text{old}} \cap C_{\text{new}} = \varnothing$. Let $K_{\text{old}} = |C_{\text{old}}|$ denote the number of old classes and $K_{\text{new}} = |C_{\text{new}}|$ the number of novel classes. The total number of categories is then $K = K_{\text{old}} + K_{\text{new}} = |Y_u|$. Following prior work, we assume that this total number of categories is known a priori~\cite{fini2021unified, han2021autonovel, zhong2021openmix, zhao2021novel}.

\subsection{Parametric GCD method}
\label{Parametric GCD method (SimGCD)}

We briefly review the original SimGCD framework~\cite{wen2023parametric}, since MedXplore is built on its parametric GCD formulation. In the original paper, SimGCD integrates \emph{representation learning} and \emph{parametric classification} with a ViT\mbox{-}B/16~\cite{dosovitskiy2020image} encoder pretrained by DINO~\cite{caron2021emerging} on ImageNet~\cite{deng2009imagenet}. This architectural detail is included only to summarize the original SimGCD design and should not be interpreted as the backbone used in our experiments; our actual experimental configuration is specified separately in Sec.~\ref{Experiment Setup}.

\noindent\textbf{Representation Learning.}
The framework combines supervised contrastive learning on labeled data and unsupervised contrastive learning on all samples. Given two augmented views $x_i$ and $x'_i$ of the same image in a mini\mbox{-}batch $B$, with $z_i=g(f(x_i))$ and $\tilde z_i=g(f(x'_i))$ where $f$ is the encoder and $g$ is an MLP projection head, the unsupervised contrastive loss is
\begin{equation}
\mathcal{L}^{u}_{\mathrm{rep}}
= -\frac{1}{|B|}\sum_{i\in B}
\log
\frac{\exp\!\left(z_i^\top \tilde z_i / \tau_u\right)}
{\sum_{n\in B}\exp\!\left(z_i^\top \tilde z_n / \tau_u\right)} .
\label{eq:1}
\end{equation}
For the labeled subset $B_\ell \subset B$, let $N_i$ be the indices sharing the same label as $x_i$. The supervised contrastive loss is
\begin{equation}
\mathcal{L}^{s}_{\mathrm{rep}}
= -\frac{1}{|B_\ell|}\sum_{i\in B_\ell}\frac{1}{|N_i|}
\sum_{q\in N_i}
\log
\frac{\exp\!\left(z_i^\top \tilde z_q / \tau_c\right)}
{\sum_{n\in B}\exp\!\left(z_i^\top \tilde z_n / \tau_c\right)} .
\label{eq:2}
\end{equation}
The total representation objective is defined as
\begin{equation}
\mathcal{L}_{\mathrm{rep}}=(1-\lambda)\,\mathcal{L}^{u}_{\mathrm{rep}}+\lambda\,\mathcal{L}^{s}_{\mathrm{rep}}.
\end{equation}

\textbf{Parametric Classification.}
SimGCD replaces $k$-means with learnable prototypes $C=\{c_1,\dots,c_K\}$, where $K=|Y_l\cup Y_u|$. For feature $h_i=f(x_i)$, the class probability is
\begin{equation}
p_i(k)=
\frac{\exp\!\left((h_i/\|h_i\|)^\top (c_k/\|c_k\|)/\tau_s\right)}
{\sum_{k'} \exp\!\left((h_i/\|h_i\|)^\top (c_{k'}/\|c_{k'}\|)/\tau_s\right)} .
\label{eq:3}
\end{equation}
Let the companion view yield a pseudo\mbox{-}label $q'_i$, and define the cross\mbox{-}entropy $\ell(q,p)=-\sum_k q(k)\log p(k)$. The unsupervised and supervised classification losses are
\begin{equation}
\mathcal{L}^{u}_{\mathrm{cls}}=\frac{1}{|B|}\sum_{i\in B}\ell(q'_i,p_i),
\;
\mathcal{L}^{s}_{\mathrm{cls}}=\frac{1}{|B_\ell|}\sum_{i\in B_\ell}\ell(y_i,p_i).
\label{eq:4}
\end{equation}
For regularization, we minimize the entropy
\begin{equation}
H(\bar{p}) = -\sum_k \bar{p}(k) \log \bar{p}(k),
\end{equation}
of the batch-mean prediction
\begin{equation}
\bar{p} = \frac{1}{2|B|}\sum_{i \in B}(p_i + p'_i).
\end{equation}
The classification objective is
\begin{equation}
\mathcal{L}_{\mathrm{cls}}
= (1-\lambda)\big(\mathcal{L}^{u}_{\mathrm{cls}}-\epsilon H(\bar p)\big)
+ \lambda\,\mathcal{L}^{s}_{\mathrm{cls}} .
\label{eq:5}
\end{equation}
Finally, the overall objective is
\begin{equation}
\mathcal{L}=\mathcal{L}_{\mathrm{rep}}+\mathcal{L}_{\mathrm{cls}}.
\end{equation}

\begin{figure*}[!t] 
    \centering 
    \includegraphics[width=\linewidth]{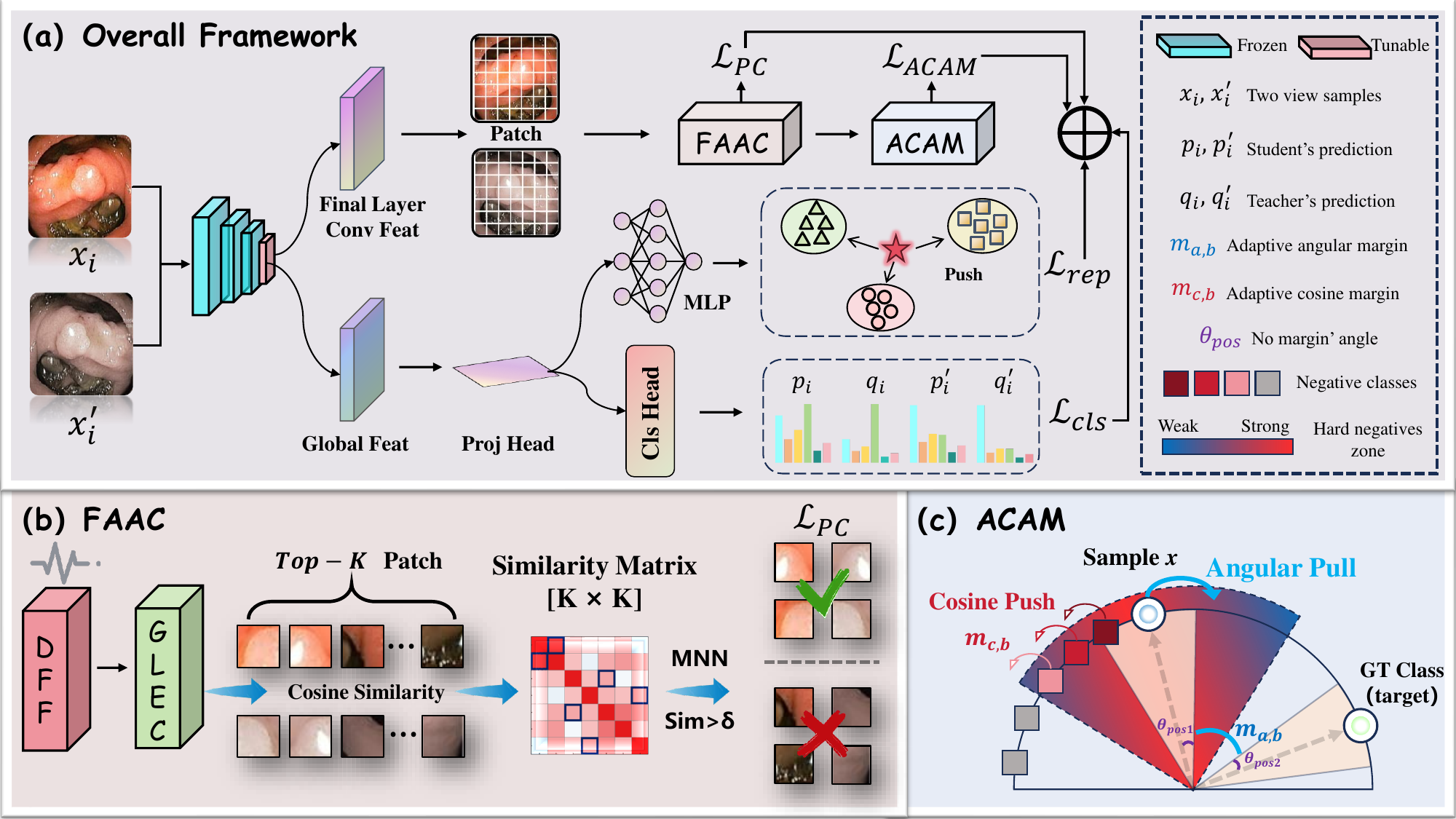}    
    \caption{\textbf{(a)} Overview of our MedXplore framework. MedXplore is mainly composed of representation learning, parametric classification, the perception branch, and ACAM.
    \textbf{(b)} The perception branch suppresses background regularities, extracts reliable semantic anchors, and supports patch consistency learning. The implementations of Dynamic Frequency Filtering (DFF) and GLEC are illustrated in Appendix.
    \textbf{(c)} ACAM dynamically adjusts margins based on semantic difficulty and feature confidence.} 
    \Description{A three-part overview of the MedXplore framework. Part (a) presents the full pipeline with representation learning and parametric classification. Part (b) zooms in on the perception branch that extracts informative patches and enforces patch consistency. Part (c) illustrates the ACAM module that adaptively adjusts angular and cosine margins according to sample confidence.}
    \label{pipline} 
\end{figure*}

\section{Our Method}
\label{sec:Method}

In this section, we present MedXplore, a medical GCD framework that addresses two challenges: unreliable lesion perception and ambiguous category boundaries. FAAC (Sec.~\ref{sec:FAAC}) improves lesion-aware representation learning through full-spectrum filtering, global-local energy contrast activation, and patch consistency. ACAM (Sec.~\ref{sec:ACAM}) then uses the resulting semantic signals to adaptively calibrate decision boundaries. Together, they provide complementary perceptual and discriminative modeling for reliable novel-class discovery. Fig.~\ref{pipline} shows the overall pipeline, and pseudo-code is provided in the Appendix.

\begin{figure}[t]
    \centering
    \includegraphics[width=\linewidth]{Figure/Filtering.pdf}
    \caption{Visual comparison of lesion activation maps under different filtering strategies. Spatial variance is distracted by confounds, PFESA~\cite{li2025pfesa} and FANet~\cite{liu2026frequency} amplify reflections and blur lesion responses. In contrast, FAAC produces more reliable semantic anchors.}
    \label{fig:filtering_dilemma}
\end{figure}

\subsection{Frequency-SNR Adaptive Attention and Consistency}
\label{sec:FAAC}

As discussed in Sec.~\ref{sec:intro}, conventional spatial attention in medical image analysis easily overfits to familiar anatomical textures of known classes, leading to severe old-class bias during novel category discovery~\cite{das2025medgcd, feng2025neighbor}. To break this spatial limitation, we shift our focus to the frequency domain, which is inherently better at capturing subtle, texture-agnostic lesion deviations~\cite{zhang2026decoding, deng2025fmnet}. However, traditional frequency processing falls short in our scenario: as shown in Fig.~\ref{fig:filtering_dilemma}, direct spatial variance is easily dominated by confounds like UI borders, while PFESA and FANet either amplify reflection noise or blur crucial lesion boundaries~\cite{li2025pfesa, liu2026frequency}. To this end, we propose FAAC, a dynamic module that leverages learnable full-spectrum filtering and local energy contrast to precisely localize lesion evidence without relying on fixed spectral bands.

\textbf{Dynamic frequency filtering.}
Given an input image \(x\), we generate two augmented views \(x^{(1)}\) and \(x^{(2)}\). The backbone extracts a feature map \(X^{(v)}\in\mathbb{R}^{B\times C\times H\times W}\) and its aligned patch tokens \(T^{(v)}\in\mathbb{R}^{B\times N\times D}\) for each view \(v\in\{1,2\}\), where \(B\) is the batch size, \(N\) is the number of patches, and \(D\) is the token dimension. To suppress repetitive anatomical background while retaining subtle lesion boundaries, we first project the spatial feature map into the frequency domain. Rather than using a fixed filter~\cite{li2025pfesa}, we separately process the real and imaginary components by depth-wise convolutions, allowing each channel to learn its own frequency response:
\begin{equation}\label{eq:freq_filter}
\begin{aligned}
\hat{X}^{(v)}_{R} &= \mathrm{DWConv}(\mathrm{Re}(\mathcal{F}(X^{(v)}))), \\
\hat{X}^{(v)}_{I} &= \mathrm{DWConv}(\mathrm{Im}(\mathcal{F}(X^{(v)}))).
\end{aligned}
\end{equation}
where \(\mathcal{F}(\cdot)\) denotes the 2D FFT, and \(\mathrm{DWConv}(\cdot)\) is a channel-wise depth-wise convolution acting as a learnable full-spectrum frequency filter.

\begin{equation}\label{eq:freq_reconstruction}
X_{\mathrm{freq}}^{(v)} = \mathrm{Re}\!\left(\mathcal{F}^{-1}\!\left(\hat{X}^{(v)}_{R} + j\,\hat{X}^{(v)}_{I}\right)\right),
\end{equation}
where \(j=\sqrt{-1}\). This reconstruction yields a frequency-enhanced feature map whose repetitive background patterns are attenuated while task-relevant spectral components are preserved, including low-frequency structural variations and high-frequency boundary cues.

\textbf{Global-local energy contrast activation.}
After frequency purification, we explicitly activate abnormal regions instead of predicting patch scores with another learned attention head. We first collapse the filtered feature map along the channel axis to obtain a spatial energy map:
\begin{equation}\label{eq:energy_map}
E^{(v)} = \frac{1}{C}\sum_{c=1}^{C} X_{\mathrm{freq},c}^{(v)} \in \mathbb{R}^{B\times 1\times H\times W}.
\end{equation}
For each location \((i,j)\), let \(\Omega_{ij}\) denote a \(3\times 3\) neighborhood and \(\bar{E}_{ij}^{(v)}\) its local mean. We compute the local energy fluctuation by
\begin{equation}\label{eq:local_variance}
V_{\mathrm{loc}}^{(v)}(i,j)=\mathrm{Var}_{(p,q)\in\Omega_{ij}}\,E^{(v)}(p,q).
\end{equation}

Guided by our dynamic full-spectrum filtering, high values of $V_{\mathrm{loc}}^{(v)}$ highlight regions that deviate structurally from their immediate surroundings. Because we do not rely on fixed spectral bands, these local deviations can flexibly capture a wide range of lesion evidence, from subtle density shifts to sharp boundary anomalies. To further ensure these signals are not hijacked by common image-level artifacts like UI borders or reflections—as traditional methods often are—we propose a Global-Local Energy Contrast (GLEC) activation. By normalizing the local variance against global spatial statistics, GLEC explicitly suppresses ubiquitous background noise and isolates genuinely abnormal lesion regions:

\begin{equation}\label{eq:glec_attention}
A^{(v)}=\sigma\!\left(\gamma\cdot
\frac{V_{\mathrm{loc}}^{(v)}-\mu(V_{\mathrm{loc}}^{(v)})}
{\mathrm{Std}(V_{\mathrm{loc}}^{(v)})+\varepsilon}\right)
,
\end{equation}
where \(\sigma(\cdot)\) is the sigmoid function, \(\mu(\cdot)\) and \(\mathrm{Std}(\cdot)\) are the global mean and standard deviation over the full spatial map, \(\gamma\) is a learnable scaling factor, and \(\varepsilon\) is a small constant for numerical stability. Unlike conventional attention heads that are easily biased toward labeled patterns, Eq.~\eqref{eq:glec_attention} highlights regions whose local energy deviates most strongly from the global context in the filtered feature space.

\begin{equation}\label{eq:weighted_feature}
X_{\mathrm{out}}^{(v)} = X^{(v)} \odot A^{(v)} + X^{(v)},
\end{equation}
where \(A^{(v)}\) is broadcast along the channel dimension. We then partition \(A^{(v)}\) with the same patch grid used to form \(T^{(v)}\), and obtain patch-level attention scores by region-wise mean pooling:
\begin{equation}\label{eq:patch_scores}
a_{b,n}^{(v)} = \frac{1}{|\Omega_n|}\sum_{(i,j)\in\Omega_n} A_{b}^{(v)}(i,j),\qquad
\mathcal{I}^{(v)} = \mathrm{TopK}(a^{(v)},K),
\end{equation}
where \(\Omega_n\) is the spatial support of the \(n\)-th patch. Using \(\mathcal{I}^{(v)}\), we gather the corresponding token features from \(T^{(v)}\) and denote the candidate sets by \(C^{(v)}\in\mathbb{R}^{B\times K\times D}\).

\textbf{Patch consistency from semantic anchors.}
Although FAAC yields stable candidate patches, common medical image augmentations, such as color jitter, cropping, and flipping, still introduce photometric and geometric perturbations that can displace corresponding regions across views~\cite{fu2023oif, desai2021vortex, meyer2021contrast}. We therefore enforce consistency on the Top-K semantic anchors selected by \(a^{(v)}\). Let \(\mathcal{S}_{b,i,j}=\cos(C_{b,i}^{(1)}, C_{b,j}^{(2)})\) be the cosine similarity between two candidate patches. We retain only mutual nearest-neighbor matches whose similarity exceeds a threshold \(\delta\):
\begin{equation}\label{eq:mutual_nn}
\mathcal{P}_{b} =
\left\{
(i,j)\ \middle|\
\begin{aligned}
&j=\arg\max_{j'\in[K]}\mathcal{S}_{b,i,j'},\\
&i=\arg\max_{i'\in[K]}\mathcal{S}_{b,i',j},\ \mathcal{S}_{b,i,j}\ge \delta
\end{aligned}
\right\},
\end{equation}
where \([K]=\{1,\ldots,K\}\) indexes the Top-K candidate pool.

For each valid match \((i,j)\in\mathcal{P}_b\), we define the pair weight as the average response of the two selected anchors,
\[
w_{b,i,j} =
\tfrac{1}{2}
\left(
a^{(1)}_{b,\mathcal{I}^{(1)}_{b,i}} +
a^{(2)}_{b,\mathcal{I}^{(2)}_{b,j}}
\right).
\]
The patch consistency loss is
\begin{equation}\label{eq:pc_loss}
\mathcal{L}_{\mathrm{pc}}=
\dfrac{1}{B'}\sum_{b\in\mathcal{B}^{*}}
\dfrac{\sum_{(i,j)\in\mathcal{P}_b} w_{b,i,j}\big(1-\mathcal{S}_{b,i,j})}
{\sum_{(i,j)\in\mathcal{P}_b} w_{b,i,j} + \varepsilon},
\end{equation}
where \(\mathcal{B}^*=\{b\mid |\mathcal{P}_b|\ge\text{min matches}\}\) is the set of valid samples, \(B'=|\mathcal{B}^*|\), and \(\varepsilon\) is a small constant for numerical stability. It is worth noting that FAAC serves exclusively as a training-time auxiliary module to guide the backbone in learning texture-agnostic, lesion-sensitive representations. FAAC is entirely discarded during inference. Consequently, MedXplore retains the exact same architectural simplicity and inference speed as the baseline model, yet benefits from the debiased representations cultivated during training.

\begin{figure}[!t]
    \centering
    \includegraphics[width=\linewidth]{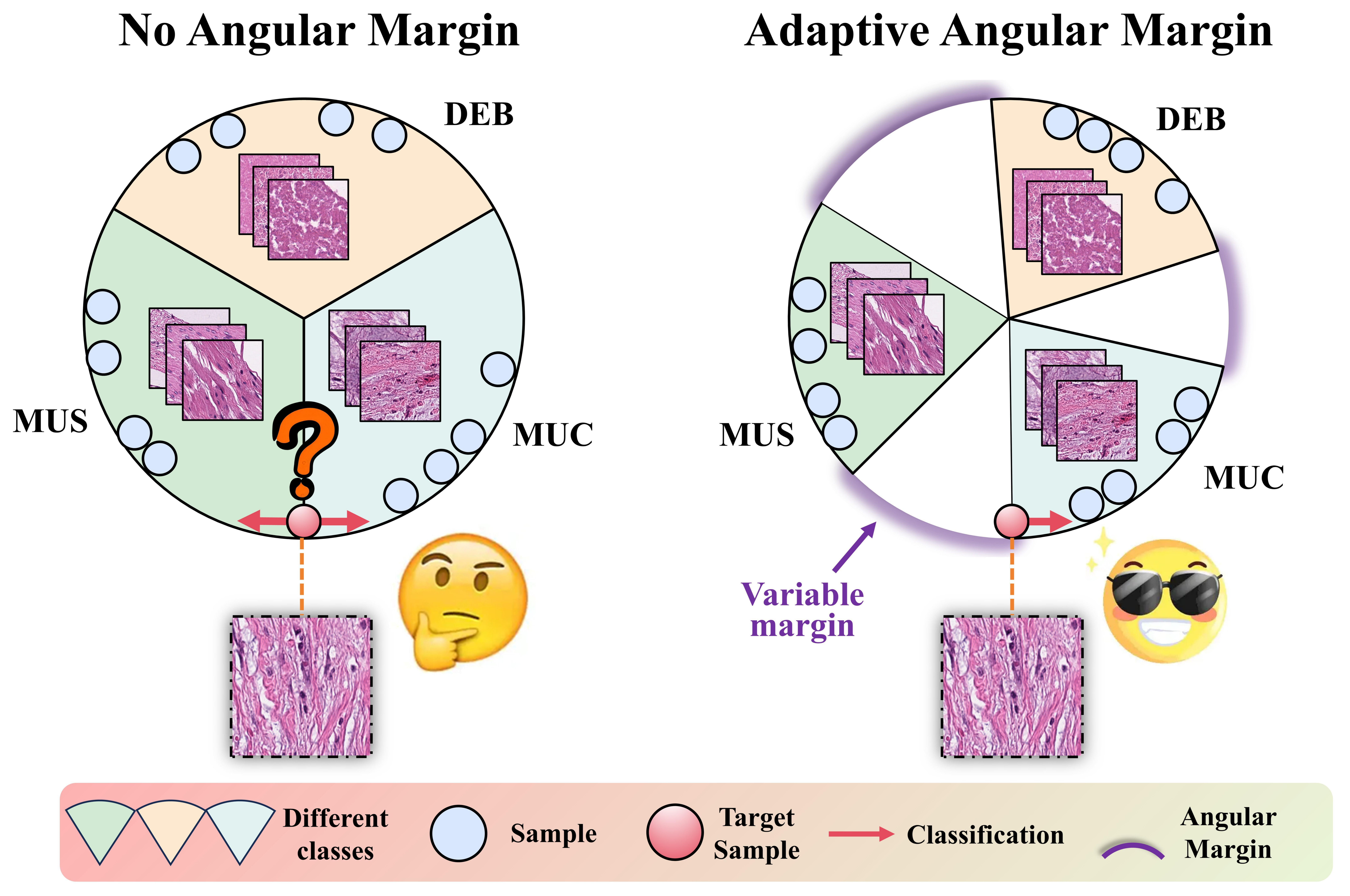}
    \caption{Conceptual illustration of how our proposed Adaptive Angular Margin resolves the classification ambiguity of boundary samples that occurs in standard models with No Angular Margin.}
    \Description{A schematic comparison of decision boundaries with and without the proposed adaptive angular margin, showing that the adaptive margin pushes ambiguous boundary samples toward a clearer separation.}
    \label{Fig:margin}
\end{figure}

\subsection{Adaptive Cosine-Angle Margin}
\label{sec:ACAM}
As discussed in Sec.~\ref{sec:intro}, medical images exhibit high heterogeneity and weak semantic margins, which often lead to cluster drift and unstable discrimination. Inspired by ArcFace~\cite{deng2019arcface} and CosFace~\cite{wang2018cosface}, we propose \textbf{ACAM}. Unlike fixed-margin formulations, ACAM uses FAAC-derived semantic anchors to adapt the margin strength to sample reliability, avoiding both over-penalization of easy samples and under-constraint of ambiguous boundary cases. As illustrated in Fig.~\ref{Fig:margin}, ACAM pushes reliable positives toward clearer angular separation while selectively suppressing nearby hard negatives. Specifically, we employ a weight-normalized linear layer as the classifier head $f_{\theta}: \mathbb{R}^D \to \mathbb{R}^C$, so that its output logits have a cosine form:
\begin{equation}\label{z_bc}
z_{b,c} =  \frac{\mathbf{w}_c \cdot \phi(x_b)}{\|\mathbf{w}_c\| \|\phi(x_b)\|} = \cos\theta_{b,c} \in [-1,1],
\end{equation}
where $\phi(x_b)$ is the feature extracted by the backbone, $\mathbf{w}_c$ is the weight vector of class $c$. For a labeled sample $x_b$ with ground-truth class $y_b$, its positive-class cosine logit is $\cos\theta_{\text{pos},b} = z_{b,y_b}$, and the maximum of all negative-class cosine logits is $\cos\theta_{\text{neg-max},b} = \max_{j \neq y_b} z_{b,j}$. The positive-negative gap is defined as $\Delta_b = z_{b,y_b} - \max_{j \neq y_b} z_{b,j}$.

\textbf{Fusion of three-dimensional confidence signals.}
To estimate sample reliability, we fuse three signals into a single score: attention confidence $A_b$ for \textit{semantic saliency}, weighted-matching confidence $W_b$ for \textit{cross-patch consistency}, and gap-based confidence $G_b$ for \textit{class separability}. After percentile normalization, they are averaged to obtain the final sample-level confidence:
\begin{equation}\label{Ab}
A_b = \frac{1}{2}\left( \frac{1}{K}\sum_{i \in \mathcal{I}_b^{(1)}} a_{b,i}^{(1)} + \frac{1}{K}\sum_{j \in \mathcal{I}_b^{(2)}} a_{b,j}^{(2)} \right),
\end{equation}

\begin{equation}\label{u_b}
u_b = \frac{1}{3}\Big(\Pi(A_b) + \Pi(W_b) + G_b\Big),
\end{equation}
where $u_b \in [0,1]$ indicates sample reliability. Here, $A_b$ is the mean attention weight of the top-$K$ patches from the two views, $W_b$ is the mean cosine similarity of the mutually matched patch pairs, and $G_b = \Pi(\Delta_b)$ is the percentile-normalized separability confidence derived from the positive-negative gap.

\textbf{Adaptive angular and cosine margins.}
Based on the confidence $u_b$, we dynamically generate an \textit{angular margin} $m_{a,b}$ applied to positive classes and a \textit{cosine margin} $m_{c,b}$ applied to negative class suppression. These adaptive margins are defined as:
\begin{equation}\label{m}
m_{a,b} = M_a^{\max} \cdot u_b,\quad m_{c,b} = M_c^{\max} \cdot u_b,
\end{equation}
where $M_a^{\max}$ and $M_c^{\max}$ denote the maximum angular margin and cosine suppression strength, respectively. The linear mapping ensures monotonicity and interpretability, allowing $u_b$ to directly control the margin strength $m$.

For the positive class $y_b$, we transform its cosine logit $\cos\theta_{\text{pos},b}$ into $\cos(\theta_{\text{pos},b} + m_{a,b})$. For all negative classes, we introduce \textit{threshold gating}, applying adaptive suppression only to ``hard negatives'' that are close to the positive class. In this way, ACAM realizes the behavior sketched in Fig.~\ref{Fig:margin}: ambiguous samples near the decision boundary receive stronger geometry-aware correction, whereas already well-separated samples are not over-constrained. We define the margin-adjusted class logit as
\begin{equation}\label{eq:acam_logit}
l_{b,c} =
\begin{cases}
s \cdot \big(\cos\theta_{b,y_b}\cos m_{a,b} - \sin\theta_{b,y_b}\sin m_{a,b}\big), & c = y_b,\\[3pt]
s \cdot \left( z_{b,c} - \dfrac{m_{c,b}}{1-\delta^{'}} \cdot \max(0, z_{b,c} - \delta^{'}) \right), & c \neq y_b,
\end{cases}
\end{equation}
where $\delta^{'} \in [0,1)$ denotes the gating threshold, $s$ is a unified scaling factor, and $z_{b,c}$ is the original cosine logit for class $c$ and sample $b$. The \textbf{ACAM loss} is then
\begin{equation}\label{acam}
\mathcal{L}_{\text{ACAM}} = -\frac{1}{|\mathcal{B}_{\text{l}}|} \sum_{b \in \mathcal{B}_{\text{l}}} \log\frac{\exp(l_{b,y_b})}{\sum_{c=1}^{C}\exp(l_{b,c})}.
\end{equation}

We restrict ACAM to labeled samples, since applying adaptive margins to pseudo-labeled unlabeled samples can amplify confirmation bias and degrade novel-class discovery. This mechanism imposes stronger discriminative constraints on high-confidence samples while applying gentler updates to low-confidence ones, thereby adaptively \textbf{balancing intra-class compactness and inter-class separability} and reducing the tendency to absorb ambiguous novel samples into dominant old classes. Since ACAM only modulates the training objective, it also introduces no additional branch at inference time. Finally, we combine the ACAM loss with the SimGCD base loss and the patch-consistency loss from FAAC as follows:
\begin{equation}\label{L}
\mathcal{L} = \mathcal{L}_{\text{rep}} + \mathcal{L}_{\text{cls}} + \alpha \cdot \mathcal{L}_{\text{pc}} + \beta \cdot  \mathcal{L}_{\text{ACAM}},
\end{equation}
where $\alpha$ and $\beta$ are balancing coefficients. Through the synergistic interplay of ACAM and FAAC, MedXplore provides complementary perceptual and discriminative modeling, thereby enabling reliable novel-class discovery.

\begin{table*}[t]
\centering
\caption{Comparative results on four medical imaging datasets. \textbf{Bold} represents the best results. $^*$Results are taken from NGUF~\cite{feng2025neighbor}.}
\label{tab:experimental_results}

\begin{tabular}{lccccccccccccc}
\toprule
\multirow{2}{*}{\centering Method} & \multicolumn{3}{c}{Kvasir} & \multicolumn{3}{c}{OrganAMNIST} & \multicolumn{3}{c}{OrganCMNIST} & \multicolumn{3}{c}{NCT-CRC-HE-100K} \\
\cmidrule(lr){2-4} \cmidrule(lr){5-7} \cmidrule(lr){8-10} \cmidrule(lr){11-13}
& All & Old & New & All & Old & New & All & Old & New & All & Old & New \\

\midrule
RS+$^*$~\cite{han2021autonovel} & 31.3 & 24.9 & 31.4 & 70.2 & 71.5 & 70.3 & 65.6 & 83.7 & 61.5 & 72.1 & 92.4 & 64.6 \\
UNO+$^*$~\cite{fini2021unified} & 24.6 & 22.6 & 21.5 & 67.3 & 71.7 & 66.4 & 64.8 & 82.3 & 60.8 & 75.4 & 95.3 & 66.2 \\
ORCA~\cite{cao2021open} & 31.5 & 31.2 & 27.3 & 69.2 & 76.0 & 68.4 & 66.1 & 86.2 & 61.6 & 70.8 & 95.5 & 57.1 \\

GCD$^*$~\cite{vaze2022generalized} & 27.4 & 25.6 & 28.8 & 75.0 & 88.3 & 72.3 & 65.7 & 81.9 & 62.0 & 76.1 & 97.6 & 63.3 \\
DCCL$^*$~\cite{pu2023dynamic} & 30.1 & 29.8 & 30.2 & 76.8 & 88.4 & 76.3 & 68.1 & 87.2 & 65.4 & 77.8 & 97.6 & 67.2 \\
CMS$^*$~\cite{choi2024contrastive} & 35.2 & 35.2 & 30.9 & 78.9 & 88.8 & 77.1 & 69.4 & 87.4 & 66.8 & 79.1 & 97.7 & 73.4 \\
SimGCD~\cite{wen2023parametric} & 72.8 & 80.4 & 66.7 & 79.4 & 89.5 & 78.1 & 69.5 & 88.1 & 67.8 & 79.4 & 97.7 & 75.9 \\
LegoGCD~\cite{cao2024solving} & 75.2 & 77.1 & 73.7 & 78.2 & 81.4 & 77.6 & 71.3 & 70.5 & 71.5 & 76.9 & \textbf{99.8} & 64.1 \\
NGUF$^*$~\cite{feng2025neighbor} & 51.1 & 54.8 & 48.1 & 84.5 & 91.7 & 82.3 & 75.2 & 90.5 & 72.0 & 90.2 & 97.8 & 84.9 \\
\midrule
\rowcolor{lightblue}
\textbf{Ours }& \textbf{91.7} & \textbf{89.4} & \textbf{93.5} & \textbf{93.6} & 96.5 & \textbf{92.9} & \textbf{82.3} & \textbf{92.8} & \textbf{80.0} & \textbf{91.3} & \textbf{99.8} & \textbf{86.5} \\

\bottomrule
\end{tabular}
\end{table*}

\section{Experiments}
\label{Experiments}

\subsection{Experiment Setup}
\label{Experiment Setup}

\textbf{Datasets.} We evaluate the effectiveness of our approach on four datasets: \textit{NCT-CRC-HE-100K}~\cite{kather2019predicting}, \textit{OrganAMNIST}~\cite{yang2023medmnist}, \textit{OrganCMNIST}~\cite{yang2023medmnist}, and \textit{Kvasir}~\cite{pogorelov2017kvasir}.
These datasets encompass colorectal cancer histopathology image patches, abdominal CT-derived organ images, and gastrointestinal disease detection images.
Following the experimental protocol in~\cite{vaze2022generalized, wen2023parametric}, we designate 50\% of the categories in each dataset as seen classes according to their class-to-sample quantity distribution.
From these seen classes, 50\% of the samples are randomly selected to construct the labeled set $\mathcal{D}_l$, while the remaining samples from both seen and novel classes form the unlabeled set $\mathcal{D}_u$.
Detailed class partitions for all datasets, including the old/new split, are provided in the Appendix together with the split statistics and implementation details.

\noindent\textbf{Evaluation Metric.}
We assess the performance of our method on the unlabeled dataset $\mathcal{D}_u$ using clustering accuracy (ACC)~\cite{vaze2022generalized, pu2023dynamic}, following established practice in prior studies.
The metric is defined as
\begin{equation}
\mathrm{ACC} = \frac{1}{|\mathcal{D}_u|} \sum_{i=1}^{|\mathcal{D}_u|} \mathbf{1}\big(y_i = f^{*}(\hat{y}_i)\big),
\end{equation}
where
\begin{equation}
f^{*} = \arg\max_{f \in \mathcal{F}} \sum_{i=1}^{M} \mathbf{1}\big(y_i = f(\hat{y}_i)\big),
\end{equation}
and $\mathcal{F}$ denotes the set of all possible one-to-one mappings between predicted labels and ground-truth labels.

\noindent\textbf{Implementation details.} Following~\cite{feng2025neighbor}, we employ a ResNet-18~\cite{he2016deep} backbone pretrained on ImageNet~\cite{deng2009imagenet} and fine-tune only its last stage, from which patch tokens are extracted. This keeps our setting aligned with NGUF and avoids any architecture advantage over the ResNet-18-based baselines. The ViT\mbox{-}B/16 encoder mentioned in Sec.~\ref{Parametric GCD method (SimGCD)} is included only to summarize the original SimGCD design. We set the non-architectural hyperparameters on OrganAMNIST and directly reuse them on the other datasets, without dataset-specific retuning or access to novel-class labels. Training uses an initial learning rate of 0.1 with cosine decay, batch size 128, 200 epochs, and $\lambda=0.35$. All experiments are run on NVIDIA Tesla V100 GPUs. Notably, FAAC and ACAM are used only as training-time auxiliary mechanisms; at test time, MedXplore discards the FAAC branch and performs prediction with the same deployed backbone-and-classifier path as the baseline. Detailed test-time complexity analysis is provided in the supplementary material.

\subsection{Quantitative Results.}
\label{Quantitative Results.}

We compare the proposed approach with several representative and state-of-the-art Generalized Category Discovery methods, including RS+~\cite{han2021autonovel}, UNO+~\cite{fini2021unified}, ORCA~\cite{cao2021open}, GCD~\cite{vaze2022generalized}, DCCL~\cite{pu2023dynamic}, CMS~\cite{choi2024contrastive}, SimGCD~\cite{wen2023parametric}, and the recent medical GCD method NGUF~\cite{feng2025neighbor}. The clustering performance on the four medical imaging datasets is summarized in Tab.~\ref{tab:experimental_results}.

MedXplore consistently outperforms all previous methods on all four datasets, establishing a new \textbf{state of the art} in medical GCD. The gains are especially pronounced on Kvasir~\cite{pogorelov2017kvasir}, where the \textit{All} accuracy rises from 75.2\% (LegoGCD) to 91.7\%, while substantial improvements are also maintained on OrganAMNIST, OrganCMNIST, and NCT-CRC-HE-100K. The consistently higher \textit{New} accuracy, particularly 93.5\% on Kvasir and 92.9\% on OrganAMNIST, shows that the proposed framework improves both joint clustering and novel-category discovery. Since the Kvasir gain is still the largest among all datasets, we next analyze this dataset explicitly rather than leaving it as an aggregate result.

\begin{table}[t]
\centering
\footnotesize
\caption{Ablation study of FAAC and ACAM on OrganAMNIST~\cite{yang2023medmnist} and OrganCMNIST~\cite{yang2023medmnist}. \textbf{Bold} indicates the best results.}
\label{tab:ablation}
\renewcommand{\arraystretch}{1.12}
\setlength{\tabcolsep}{0pt}
\begin{tabular*}{\linewidth}{@{\extracolsep{\fill}}cccccccccc@{}}
\toprule
\multirow{2}{*}{ID} & \multirow{2}{*}{Baseline} & \multirow{2}{*}{FAAC} & \multirow{2}{*}{ACAM}
& \multicolumn{3}{c}{OrganAMNIST} & \multicolumn{3}{c}{OrganCMNIST} \\
\cmidrule(lr){5-7} \cmidrule(lr){8-10}
& & & & All & Old & New & All & Old & New \\
\midrule
\textbf{1} & \cmark & \xmark & \xmark & 79.4 & 89.5 & 78.1 & 69.5 & 88.1 & 67.8 \\
\textbf{2} & \cmark & \cmark & \xmark & 91.5 & 95.8 & 90.5 & 79.2 & 92.1 & 76.4 \\
\textbf{3} & \cmark & \xmark & \cmark & 89.3 & 93.2 & 88.4 & 75.9 & 91.2 & 72.5 \\
\midrule
\rowcolor{rowblue}
\textbf{4} & \cmark & \cmark & \cmark & \textbf{93.6} & \textbf{96.5} & \textbf{92.9} & \textbf{82.3} & \textbf{92.8} & \textbf{80.0} \\
\bottomrule
\end{tabular*}
\end{table}



\subsection{Qualitative Results.}
\label{Qualitative Results.}

\begin{figure}[!t]
    \centering
    \includegraphics[width=\linewidth]{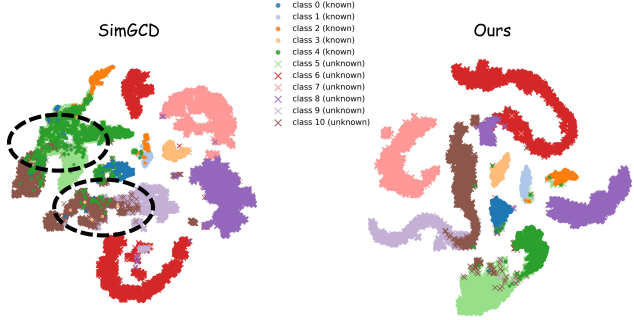}
    \caption{Visualisation of the embedding space with t-SNE on OrganAMNIST~\cite{yang2023medmnist}.}
    \Description{A t-SNE visualization comparing the embedding distributions of a baseline method and MedXplore on OrganAMNIST. MedXplore forms tighter within-class clusters and clearer separation between classes.}
    \label{tsne}
\vspace{-2mm}
\end{figure}

\textbf{Feature Visualization.} We use t-SNE to visualize the feature space on OrganAMNIST~\cite{yang2023medmnist}. As shown in Fig.~\ref{tsne}, our reliable and unbiased approach creates clearer margins and tighter clusters compared to SimGCD~\cite{wen2023parametric}, demonstrating better classification discriminability. Additional attention visualizations are already shown in Fig.~\ref{Absfig}(a) and in the supplementary material.

\begin{figure}[t]
    \centering
    \includegraphics[width=\linewidth]{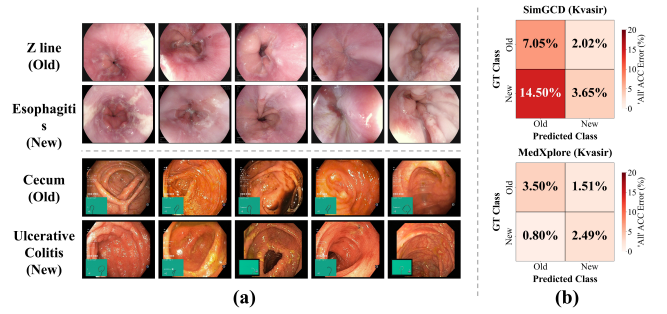}
    \caption{\textbf{Kvasir ambiguity and bias analysis.} \textbf{(a)} Representative old/new class pairs with severe visual ambiguity. \textbf{(b)} Error decomposition on Kvasir. MedXplore sharply reduces false-old errors (new$\rightarrow$old), indicating substantially alleviated old-class bias.}
    \Description{A two-part figure analyzing Kvasir. Part (a) shows representative endoscopic images from two old classes and two visually similar new classes. The paired categories share highly similar global appearance and differ mainly in subtle local lesion or mucosal patterns. Part (b) shows two 2-by-2 heatmaps comparing SimGCD and MedXplore on Kvasir. The rows correspond to ground-truth old and new groups, and the columns correspond to predicted old and new groups. SimGCD has a large new-to-old false-old error, while MedXplore greatly reduces that cross-group bias and keeps the remaining error terms low and balanced.}
    \label{fig:kvasir_bias}
\end{figure}

\begin{figure*}[ht]
    \centering
    \includegraphics[width=\linewidth]{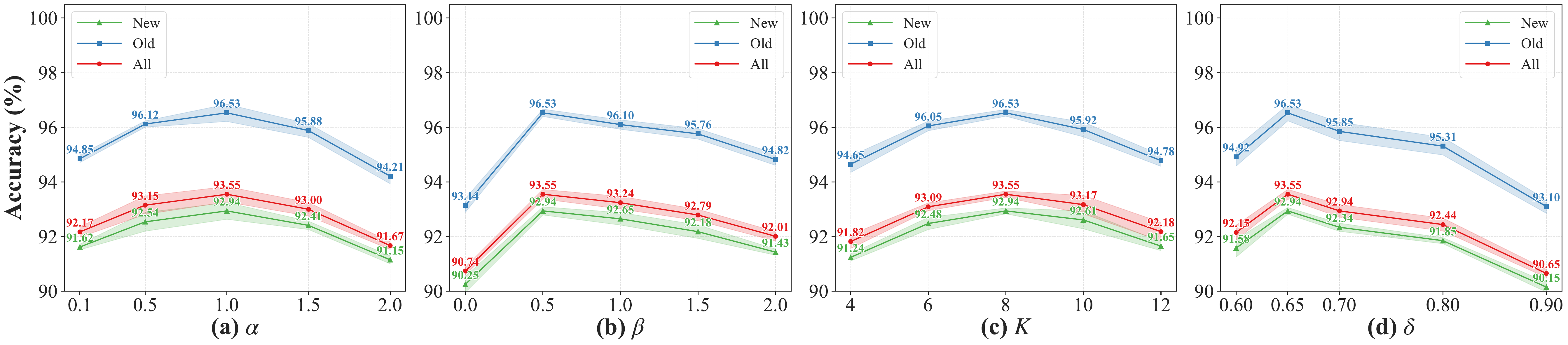}
    \caption{Effect of different weights of $\alpha$ and $\beta$ on OrganAMNIST~\cite{yang2023medmnist}. The sensitivity to $K$ values and similarity threshold $\delta$ on OrganAMNIST~\cite{yang2023medmnist}.}
    \Description{A four-panel set of line plots showing ablation sensitivity on OrganAMNIST. The plots report all, old, and new accuracy as the loss weights alpha and beta vary, and as the patch number K and similarity threshold delta vary.}
    \label{para}
\end{figure*}

\noindent \textbf{Visualization of Category Bias.} Kvasir provides a particularly informative case study because several old and novel categories are highly similar and can appear almost indistinguishable at the global level, as shown in Fig.~\ref{fig:kvasir_bias}(a). In this setting, baselines that rely on global contrastive features often fail to capture subtle structural discrepancies, which makes the old-class bias especially severe. As shown in Fig.~\ref{fig:kvasir_bias}(b), we decompose the errors on Kvasir according to the ground-truth category groups and the predicted category groups. For SimGCD, the dominant term remains the \emph{false-old} error (new$\rightarrow$old) at 14.50\%, which is still substantially larger than the other error components, namely 7.05\% true-old confusion, 2.02\% false-new error, and 3.65\% true-new confusion. MedXplore reduces the false-old term to 0.80\%, while the remaining error terms stay comparatively low at 3.50\%, 1.51\%, and 2.49\%. These results indicate that the large gain on Kvasir mainly comes from alleviating old-class bias under severe old/new ambiguity. This is mainly attributable to the design of MedXplore: FAAC focuses on discriminative lesion regions, while ACAM further separates ambiguous boundary samples, thereby substantially reducing the bias induced under ambiguous conditions.

\subsection{Ablation Study}
\label{Ablation Study}

In this section, we conduct a series of rigorous ablation studies to validate the effectiveness of our proposed FAAC and ACAM on the OrganAMNIST~\cite{yang2023medmnist} and OrganCMNIST~\cite{yang2023medmnist} datasets. The results are summarized in Tab.~\ref{tab:ablation}, while additional ablations are provided in the Appendix.

\noindent\textbf{Effectiveness of FAAC.}
Tab.~\ref{tab:ablation} isolates the system-level contributions of FAAC and ACAM. Compared with the SimGCD baseline (row 1), adding only FAAC (row 2) raises the \textit{All} accuracy from 79.4\% to 91.5\% on OrganAMNIST and from 69.5\% to 79.2\% on OrganCMNIST. This shows that FAAC provides strong representation gains by constructing robust semantic anchors before any adaptive-margin refinement is applied. Importantly, the gain does not come from hard-coding a single spectral prior: as illustrated in Fig.~\ref{fig:filtering_dilemma}, direct spatial variance is dominated by background confounds, fixed high-pass filtering over-amplifies reflections, and fixed low-pass filtering is corrupted by boundary-induced energy bleeding. FAAC avoids these failure modes through learnable full-spectrum filtering followed by context-normalized activation. Because FAAC is a sequential pipeline rather than a set of interchangeable branches, we report its internal incremental buildup separately in the supplementary material. Those results show a consistent improvement when progressively adding the frequency filter, GLEC activation, and $\mathcal{L}_{\text{PC}}$, confirming that all three components are necessary to realize the full benefit of FAAC.

\begin{table}[t]
\centering
\footnotesize
\caption{Incremental internal ablation of FAAC on top of the SimGCD baseline. ACAM is disabled in all rows.}
\label{tab:faac_seq_ablation}
\renewcommand{\arraystretch}{1.1}
\setlength{\tabcolsep}{3.0pt}
\begin{tabular}{@{}c c c c *{6}{c}@{}}
\toprule
\multirow{2}{*}{ID} & \multicolumn{3}{c}{FAAC} & \multicolumn{3}{c}{OrganAMNIST} & \multicolumn{3}{c}{OrganCMNIST} \\
\cmidrule(lr){2-4} \cmidrule(lr){5-7} \cmidrule(lr){8-10}
& Freq. Filter & GLEC & $\mathcal{L}_{\text{PC}}$ & All & Old & New & All & Old & New \\
\midrule
\textbf{A} & \xmark & \xmark & \xmark & 79.4 & 89.5 & 78.1 & 69.5 & 88.1 & 67.8 \\
\textbf{B} & \cmark & \xmark & \xmark & 83.2 & 89.8 & 81.6 & 71.2 & 89.4 & 67.2 \\
\textbf{C} & \cmark & \cmark & \xmark & 86.4 & 91.5 & 85.2 & 73.5 & 90.5 & 69.8 \\
\midrule
\rowcolor{rowblue}
\textbf{D} & \cmark & \cmark & \cmark & \textbf{91.5} & \textbf{95.8} & \textbf{90.5} & \textbf{79.2} & \textbf{92.1} & \textbf{76.4} \\
\bottomrule
\end{tabular}
\vspace{-2mm}
\end{table}

\noindent\textbf{Effectiveness of ACAM.}
Adding only ACAM to the baseline (row 3) also substantially improves performance, reaching 89.3\% and 75.9\% \textit{All} accuracy on OrganAMNIST and OrganCMNIST, respectively. This verifies that adaptive margin optimization is effective even without the FAAC perception module. When FAAC and ACAM are combined (row 4), the model achieves the best performance on all metrics, reaching 93.6/96.5/92.9 on OrganAMNIST and 82.3/92.8/80.0 on OrganCMNIST. Relative to row 2, ACAM further improves the \textit{All} accuracy by 2.1 and 3.1 percentage points on the two datasets, while relative to row 3, FAAC contributes an additional 4.3 and 6.4 points. These results demonstrate that the two modules are complementary: FAAC improves representation quality, whereas ACAM further sharpens the decision boundaries. We provide an additional ablation over the confidence formulation $u_b$ in the supplementary material, which further supports the effectiveness of the ACAM design.

\noindent\textbf{Hyperparameter Analysis.} We present the impact of hyperparameters $\alpha$ and $\beta$ in Fig.~\ref{para}. As indicated in Eq.~\ref{L}, $\alpha$ and $\beta$ balance the contributions of the patch consistency and adaptive margin loss. First, fixing $\beta=0.5$, increasing $\alpha$ from 0.1 to 1.0 improves the \textit{All}/\textit{Old}/\textit{New} accuracies from 92.2/94.9/91.6 to 93.6/96.5/92.9, showing that patch consistency is crucial for stabilizing cross-view semantic anchors. When $\alpha$ is further increased to 1.5 and 2.0, all three metrics drop, suggesting over-regularization. Second, fixing $\alpha=1.0$, the best performance is achieved at $\beta=0.5$, while larger $\beta$ values gradually degrade both known-class compactness and novel-class discovery. Overall, $\alpha=1.0$ and $\beta=0.5$ provide the best balance.

\noindent\textbf{Impact of Patch Number $K$.} We compare accuracy with different $K$ for FAAC in Fig.~\ref{para} (c). Increasing $K$ from 4 to 8 consistently improves performance, with the \textit{All}, \textit{Old}, and \textit{New} accuracies reaching 93.6\%, 96.5\%, and 92.9\% at $K=8$. When $K$ is further increased to 10 and 12, all three metrics decline, indicating that including too many patches introduces less informative regions and weakens the semantic-anchor quality. Therefore, we select $K=8$ as the best tradeoff between representation completeness and patch selectivity.

\noindent\textbf{Impact of Similarity Threshold $\delta$.} We compare accuracy with different similarity thresholds $\delta$ for the patch consistency in Fig.~\ref{para} (d). The best performance is achieved at $\delta=0.65$, where the \textit{All}, \textit{Old}, and \textit{New} accuracies reach 93.6\%, 96.5\%, and 92.9\%, respectively. A smaller threshold such as 0.60 admits more noisy matches, while stricter thresholds from 0.70 to 0.90 remove too many valid cross-view correspondences and gradually degrade performance. This result indicates that $\delta=0.65$ provides the most reliable balance between match quality and match coverage.

\section{Conclusion}
\label{sec:conclusion}

Current GCD methods struggle to generalize in medical imaging due to domain-specific biases, subtle lesion morphology, and substantial modality variation. We propose MedXplore, a unified framework for unbiased and reliable generalized category discovery in complex medical scenarios. Through the integration of FAAC and ACAM, MedXplore jointly models perceptual and discriminative cues, enabling frequency-aware semantic anchor extraction and adaptive refinement of class boundaries. By relying on learnable full-spectrum filtering rather than static high-pass or low-pass assumptions, this design mitigates feature bias, stabilizes representations, and supports unbiased discovery of novel disease categories. Extensive experiments on multiple medical datasets demonstrate that MedXplore achieves SOTA performance, and the dedicated Kvasir analysis further shows that it sharply reduces false-old errors under severe old-new ambiguity. At the same time, the current benchmarks mainly cover lesions with relatively localized discriminative cues; evaluating robustness on diffuse low-frequency pathologies remains an important direction for future work.

\FloatBarrier
\bibliographystyle{ACM-Reference-Format}
\bibliography{main}

\end{document}